\documentclass[10pt,letterpaper]{article}
\usepackage{bbm}
\usepackage{mathrsfs}
\usepackage{amsfonts}
\usepackage{opex3}
\usepackage{amssymb}
\usepackage{graphicx}
\usepackage{cite}
\usepackage[linesnumbered,ruled,lined]{algorithm2e}
\usepackage{array}
\usepackage{subfigure}
\usepackage{multirow}
\usepackage{color}
\usepackage{amsmath}
\usepackage{bm}
\usepackage{upgreek}
\usepackage{float}
\newcommand{\format}[1]{\text{\bf #1}}
\newcommand{\FORMAT}[1]{\text{\bf #1}}

\usepackage{fmtcount}
\usepackage{booktabs,threeparttable}

\begin{document}

\title{Multi-frame denoising of high speed optical coherence tomography data using inter-frame and intra-frame priors}

\author{Liheng Bian, Jinli Suo, Feng Chen and Qionghai Dai$^*$}

\address{Department of Automation, Tsinghua University, Beijing, 100084, China}

\email{qhdai@mail.tsinghua.edu.cn} 



\begin{abstract}
Optical coherence tomography (OCT) is an important interferometric diagnostic technique which provides cross-sectional views of the subsurface microstructure of biological tissues. However, the imaging quality of high-speed OCT is limited due to the large speckle noise. To address this problem, this paper proposes a multi-frame algorithmic method to denoise OCT volume. Mathematically, we build an optimization model which forces the temporally registered frames to be low rank, and the gradient in each frame to be sparse, under logarithmic image formation and noise variance constraints.
Besides, a convex optimization algorithm based on the augmented Lagrangian method is derived to solve the above model. The results reveal that our approach outperforms the other methods in terms of both speckle noise suppression and crucial detail preservation.
\end{abstract}


\bibliographystyle{osajnl_suo}
\bibliography{OCT_Denoise}


\section{Introduction}

Optical coherence tomography (OCT), which dates back to 1991 \cite{First1991}, provides cross-sectional views of the subsurface microstructure of biological tissues \cite{Overview1999, Overview2001}, and has become a widely used diagnostic technique in the medical field due to its non-invasive nature. For example, ophthalmology has particularly benefited from OCT, as it can image the retina and aid in the diagnosis \cite{Multiframe_Wavelet}. Different from common CCD imaging, OCT is an interferometric technique, and typically uses the near-infrared laser light to penetrate into the scattering medium before capturing the backscattered optical waves for the final imaging \cite{SpeckleInOCT}. With the development of ultrahigh resolution OCT (UHROCT) \cite{UHROCT} and Fourier domain OCT (FDOCT) \cite{FDOCT_1,FDOCT_2}, it is feasible to visualize biological tissues at a cellular level, and up to the depth of nearly 1mm below the surface with high sensitivity and image quality. Furthermore, the image acquisition speed of OCT systems has greatly improved with the development of high speed sensors and tunable lasers with MHz scanning rate, which allows real-time imaging of vivo tissues \cite{FDOCT_2}.

One of the important challenges limiting high-speed OCT's development is its unsatisfying image quality caused by speckle noise. Due to the coherence of optical waves, speckle noise arises under limited spatial-frequency bandwidth of the interference signals \cite{SpeckleInOCT}. The generation mechanism of OCT determines that the properties of speckle noise are related not only to the laser source but also to the tissue's structural properties \cite{SpeckleModel_1, SpeckleModel_2, SpeckleModel_3}, and thus results in non-uniform speckle noise over the entire image. Due to the significance of high precision in medical diagnosis, it is vital to remove speckle noise from OCT images for image quality enhancement.

Large efforts have been taken for denoising OCT images, and various approaches are reported. These methods mainly fall into two categories, namely {\em single-frame methods} and {\em multi-frame methods}. Single-frame methods often assume a prior model (either parametric or non-parametric) for the latent signal and noise, and then remove the noise determinatively or probabilistically from the input single image. Filtering is a widely used strategy and there are a bunch of OCT denoising filters. For example, Ozcan et al. \cite{Single_0} apply various digital filters for denoising OCT images, and the results indicate that the nonorthogonal wavelet filter together with the enhanced Lee and the adaptive Wiener filter can significantly reduce speckle noise. Based on the wavelet filter, Yue et al. \cite{Single_3} utilize the iterative edge enhancement feature of nonlinear diffusion to improve the denoising results. Similarly, \cite{Single_4} uses nonlinear diffusion in the Laplacian pyramid domain to filter ultrasonic images. The Kovesi Nw filtering technique and the Laplacian pyramid nonlinear diffusion (LPND) technique are unified together in \cite{Single_1} to remove both shot noise and speckle noise from OCT images. Besides the filtering techniques, there are also some other approaches such as regularization and Bayesian inference. The work in \cite{Single_content} uses a regularization method to minimize the Csiszar's I-divergence measurement, which would extrapolate additional details from the input noisy images to improve the visual effects. Wong et al. \cite{Bayesian} and Cameron et al. \cite{Noise_Estimation} use a statistic Bayesian least square model to reduce OCT speckle noise in logarithmic space. Xie et al. \cite{Single_2} take image contents into consideration and propose a salient structure extraction algorithm combining an adaptive speckle suppression term, to enhance ultrasound images. Different from the above filtering or statistic methods, based on sparse coding, the work in \cite{OCT_Sparse}
learns an over complete dictionary from high signal-to-noise-ratio (SNR) images, and then utilizes this dictionary to reconstruct low-SNR OCT images and achieves significant noise suppression. In all, making use of the intrinsic redundancy of a single OCT frame can help noise removal to a large extent.

Benefiting from the development of high speed OCT imaging systems, the correlation between adjacent frames increases, and researchers gradually begin to use multiple OCT frames to attenuate speckle noise and propose various multi-frame methods. Technically these methods could be further classified into hardware methods and algorithmic methods. The general idea of hardware methods is to change the parameters of OCT imaging systems to decorrelate speckle noise in different frames, and then directly average these frames after image registration to get a noise-free OCT image. For example, \cite{SpeckleModel_2} and \cite{MultiHard_3, MultiHard_4, MultiHard_5} alter the angle of incident light during capturing different frames, \cite{MultiHard_6, MultiHard_7} change the detection angle of backreflected light, and \cite{MultiHard_8} changes the frequency of the laser beam. The main disadvantage of these hardware methods is the complex procedure of data acquisition, which would greatly increase the design complexity of OCT imaging systems \cite{Noise_Estimation}. There are several recently-reported algorithmic methods for denoising multiple OCT images, making use of the redundancy of latent sharp OCT images in the frequency domain. For example, the work in \cite{MultiSoft_1} performs the 3D curvelet transform to the volume data, then thresholds the coefficients, and finally does the inverse 3D curvelet transform to realize noise removal. Under the same framework, Mayer et al. \cite{Multiframe_Wavelet} choose the wavelet domain for coefficient thresholding. In spite of the promising performance, these denoising algorithms run the risk of losing crucial details by directly truncating the coefficients.

Making use of both the intra-frame and inter-frame redundancy of OCT volume data, this paper proposes an algorithmic multi-frame optimization method to denoise OCT images. Within each single frame, an OCT image is statistically similar to a natural image and its pixel gradient map tend to be sparse. This serves as the intra-frame prior. To avoid piecewise constant artifacts by simply using the total variation constraint, and considering the excellent detail-preservation performance of the low rank prior in matrix completion \cite{Lowrank_1, Lowrank_2} and image reconstruction \cite{Lowrank_3, Lowrank_4}, we make use of the inter-frame redundancy by registering the OCT images along the image count dimension to form a low-rank volume. Subjecting to both the image formation model and the non-parametric bound constraint of non-uniform speckle noise, we build a preliminary non-convex optimization model which jointly minimizes the rank of temporally registered OCT volume and forces the sparsity of its spatial gradient.
To solve the above model, we first perform some mathematical transformations and approximations for convexification. Then considering the superior convergence property of the augmented Lagrange multiplier (ALM) method \cite{ALM} for solving constrained optimization problems, as utilized in \cite{Convex_3},
we derive a numeric algorithm based on the ALM method to solve the convexified optimization model. Experiments on both pig-eye and human retina OCT data show that our denoising technique could effectively reduce speckle noise while preserving crucial details, and exhibits superior performance compared to the other popular methods.

The remainder of this paper is organized as follows: Sec.~\ref{sec:Method} sequentially describes the preprocessing operations---frame registration, noise variance estimation, model construction, and algorithm derivation. Then in Sec.~\ref{sec:Experimental Results}, we apply our method to real-captured OCT images including pigeye and human retinal data, and compare our approach with several other popular methods in terms of both visual quality and quantitative evaluation. Finally, we conclude this paper with some conclusions and discussions in Sec.~\ref{sec:Conclusions and discussions}.
\section{Method}\label{sec:Method}

\begin{figure*}[t]
  \centering
  \includegraphics[width=\textwidth]{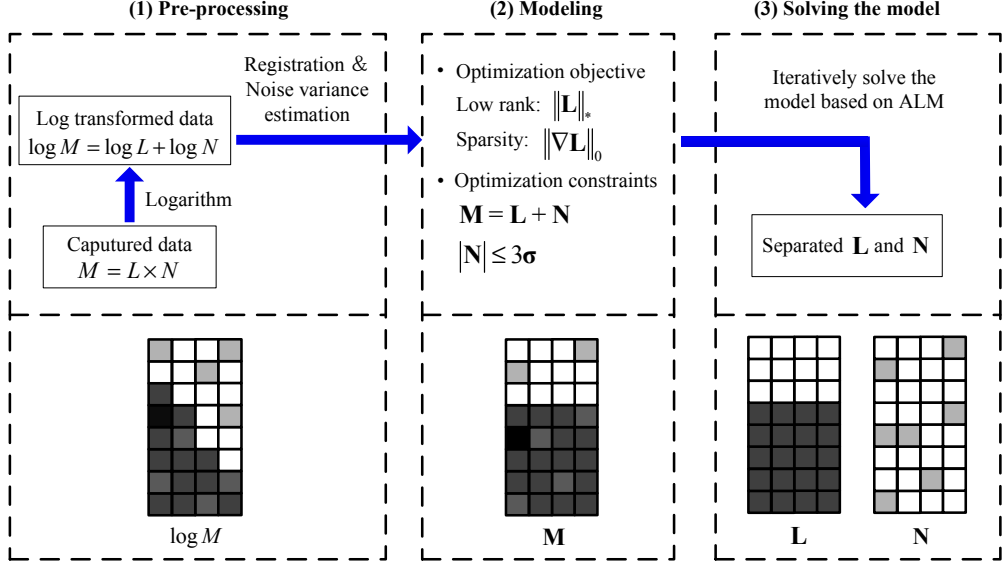}
  \caption{\textbf{Framework of the proposed approach.} After taking logarithm, each OCT frame is represented by a single column in log transformed space, and there is slight misalignment of one frame compared to other frames, as $\log M$ shows. Then by frame registration and noise variance estimation, we get not only the optimization constraints but also the optimization objective terms---the low rank of $\format L$ and the sparsity of $\bigtriangledown \format L $. Finally, the model is iteratively solved by a convex optimization algorithm based on the augmented Lagrange multiplier method, and thus $\format N$ is separated from $\format L$.}
  \label{fig:Framework}
\end{figure*}

In this section, we explain the whole operation framework of our approach as diagramed in Fig.~\ref{fig:Framework}. Our method mainly includes three steps: (1) pre-processing step including pixel logarithm, frame registration and noise variance estimation, (2) modeling step, and (3) solving step based on the ALM method. Implementation of each step is detailed in the following subsections.

\subsection{Pre-processing}\label{sec:NoiseEstimation}
In this subsection, we conduct three pre-processing operations to the captured OCT frames, including pixel logarithm, frame registration and noise variance estimation.
Due to the scattering of the laser source, the speckle noise in OCT images is multiplicative \cite{SpeckleModel_1, SpeckleModel_2}, and can be described as
\begin{equation}
M(s) = L(s)\times N(s),
\label{eqs:Log}
\end{equation}
where $M(s)$ denotes the captured data at location $s$, while $L(s)$ and $N(s)$ respectively denote the ground truth data and measurement noise at the same location. To convert the correlation between $L$ and $N$ from multiplication to addition, we take logarithm to both sides of Eq.~(\ref{eqs:Log}) and get
\begin{equation}
\log M(s) = \log L(s) + \log N(s).
\label{eqs:LogToAdd}
\end{equation}
In the following, we assume that all the variables have been logarithmically transformed.

The second preprocessing operation is frame registration. Although the optical coherence tomography facilities are usually of high capturing speed, there always tends to be slight misalignment among OCT image sequences in vivo imaging, due to the object motion and other system factors \cite{Multiframe_Wavelet}. Here we adopt the registration method in \cite{Multiframe_Wavelet}, where a powell optimizer is utilized for minimizing the sum of squared distances (SSD) among multiple registered images. Specifically, the approach applies translations and rotations to warp the pixels describing the same tissue position in different frames to the same image coordinate.

The third requisite preprocessing operation is the speckle noise variance estimation. Considering the satisfying performance of the median absolute deviation (MAD) method for noise estimation \cite{MAD}, we here utilize the MAD method as described in \cite{Noise_Estimation}. 
Due to the similar tissue properties and light directions in the same neighborhood, we assume uniform noise variance for each pixel within a small patch. Therefore, the MAD within a small neighborhood $\mathbb{N}$ of pixel $s$ is first computed in logarithmic space as \cite{Noise_Estimation}
\begin{equation}
\hat{\sigma}(s, \mathbb{N}) = 1.4826\mathscr{M}_{\mathbb{N}(s)}\left(|\log M(s_i) - \mathscr{M}_{\mathbb{N}(s)}(\log M) |\right),
\end{equation}
where $\mathscr{M}_{\mathbb{N}(s)}$ denotes the median value over $s$'s neighborhood $\mathbb{N}(s)$, and $s_i \in \mathbb{N}$ is the $i$th neighboring pixel of $s$. To make the estimated deviation more precise, we choose a larger neighborhood $\mathbb{N}_2(s)$, and calculate the local standard deviation $\hat{\sigma}$ of its sub-neighborhood $\mathbb{N}_1(s)$. Then we regard the mode of these $\hat{\sigma}$ as the preliminary noise deviation at position $s$
\begin{equation}
\overline{\sigma}(s) = mode_{\mathbb{N}_1(s)\in\mathbb{N}_2(s)}(\hat{\sigma}(s, \mathbb{N}_1)).
\end{equation}
Finally, to force the smoothness of noise variance among adjacent pixels, we conduct a cubic spline fitting process to amend $\overline{\sigma}$ and get the final standard deviation estimation of the OCT noise (corresponding noise variance can be calculated as the square of the estimated standard deviation).
Empirical studies in \cite{Noise_Estimation} tell that the noise estimation performs best when the pixel numbers of $\mathbb{N}_1$ and $\mathbb{N}_2$ are $9\times 9$ and $15\times 15$, respectively.

\subsection{Modeling}

In this subsection, we build our optimization model incorporating both of the inter-frame and intra-frame priors.
Suppose that there are $k$ frames in the OCT volume, and the pixel number of each frame is $m\times n$. We denote the temporally registered noisy OCT images, their latent sharp version, and the measurement noise as $\format M$, $\format L$ and $\format N$, respectively. Mathematically,  the image formation equation can be written as
\begin{equation}
\format M = \format L + \format N.
\label{eqs:ImageFormation}
\end{equation}

By representing each frame as a column vector, the dimensions of $\format M$, $\format L$, and $\format N$ are all $(m\times n)\times k$. After the frame registration, theoretically the entries in one specific row of $\format L$ should be exactly the same, as shown in Fig.~\ref{fig:Framework}. Therefore, we treat $\format L$ as a low rank matrix, and use the minimization of its nuclear norm $||\format L||_*$, which calculates the sum of $\format L$'s singular values \cite{Convex_3}, as the inter-frame prior constraint.

According to the statistical studies \cite{Gradient_1, Gradient_2}, the adjacent pixels in natural images own similar intensities. Thus the image gradient centers around zero and follows a heavy tailed distribution, i.e., the gradient of natural images is sparse.
Although captured via a different imaging mechanism from usual CCD imaging methods, the OCT images still follow similar statistics, and we impose the gradient sparsity of the latent OCT images as the intra-frame prior.
Specifically, the $l_0$ norm which counts the number of non-zero entries in a matrix can model the sparseness quite well, and thus we can minimize $||\nabla \format L||_0$ as the intra-frame constraint. 
Adopting the same representation in \cite{Gradient_3}, here we use matrix multiplication for gradient calculation, namely $||\nabla \format L||_0 = \sum_{a=1}^{2}||\format H_a\format L||_{0}$, where $\format H_{1}$ and $\format H_{2}$ are respectively the horizontal and vertical gradient operators, and are defined as the diagonal matrices of corresponding high pass filters $h_{1}=[-1,1]$ and $h_{2}=[-1;1]$.

As mentioned before, speckle noise includes both the image information and the zero-mean noise. Since what we concern most is the removal of the latter component, we treat the first component as a part of the latent sharp image, and concentrate on attenuating the zero-mean noise.
According to the three sigma rule, which indicates that nearly all (99.73\%) of the instances of a random variable lie within 3 times standard deviation from its mean, we can approximatively formulate the noise constraint as
\begin{equation}
|\format N| \leq 3\bm{\sigma},
\label{eqs:Noise}
\end{equation}
where $\bm{\sigma}$ is the standard deviation matrix whose dimension is $(m\times n)\times k$. By introducing a nonnegative matrix variable $\bm{\varepsilon}$, we can transform the above inequality into an equality as
\begin{equation}
\format N\odot \format N - 9 \bm{\sigma}\odot \bm{\sigma} + \bm{\varepsilon} = \format 0,
\label{eqs:Noise_2}
\end{equation}
in which $\odot$ is the entry-wise product, i.e., for two matrices $\bf{X}$ and {\bf{Y}}, $(\mathbf{X}\odot\mathbf{Y})_{ij}= \mathbf{X}_{ij}\mathbf{Y}_{ij}$.

Based on the above notations, the optimization model for denoising is defined as
\begin{eqnarray}\label{eqs:standard model_1}
\{\format L^*, \format N^*\} = \arg\min && ||\format L||_* + \lambda\sum_{a=1}^{2}||\format H_a\format L||_{0} \\ \nonumber
s.t. && \format M = \format L + \format N \\ \nonumber
&& \format N\odot \format N - 9 \bm{\sigma}\odot \bm{\sigma} + \bm{\varepsilon} = \format 0, \nonumber
\end{eqnarray}
with $\lambda$ being a positive weighting parameter to balance different objective regulation terms.

\subsection{Solving the model}
In this subsection, we derive our optimization algorithm based on the ALM method to solve the above model in (\ref{eqs:standard model_1}).
The model is obviously non-convex, so we first conduct several convexification transformations to the model.
As shown in \cite{Convex_1, Convex_2}, replacing the $l_0$ norm with the $l_1$ norm is one typical convexification transformation. Here we replace $||\format H_a\format L||_{0}$ with $||\format H_a\format L||_{1}$, where $||\cdot||_{1}$ denotes the sum of the matrix entries' absolute values.
Further, it is hard to directly utilize the ALM method to solve the convexified objective function due to its high nonlinearity. To address this problem, we replace the variables whose nuclear norm or $l_1$ norm needs minimization with two introduced auxiliary variables $\format S_1$ and $\format S_2$. Besides, we pack $\sum_{a=1}^{2}||\format H_a\format L||_1$ as $||\format P\format L||_1$ for computation simplicity, where $\format P = [\format H_1;\format H_2]$. By the above substitutions, we can rewrite the model as
\begin{eqnarray}\label{eqs:standard model_2}
min. && ||\format S_{1}||_* + \lambda ||\format S_{2}||_1 \\ \nonumber
s.t. && \format G_1 = \format S_{1} - \format L \\ \nonumber
&& \format G_2 = \format S_{2} - \format P\format L \\ \nonumber
&& \format G_3 = \format M - \format L + \format N \\ \nonumber
&& \format G_4 = \format N\odot \format N - 9 \bm{\sigma}\odot \bm{\sigma} + \bm{\varepsilon}.
\end{eqnarray}
Here $\format G_{1\dots4}$ are supposed to be $\format 0$ in theory.

As stated before, we utilize the ALM method to solve the above model. This method adopts an iterative optimization strategy, and successively updates every variable within each iteration. In the following, we derive the updating rules for each variable. First, the augmented Lagrangian function of Eq.~(\ref{eqs:standard model_2}) is
\begin{eqnarray}\label{eqs:lag_1}
f = ||\format S_{1}||_* + \lambda ||\format S_{2}||_1 + \sum_{j=1}^{4}\left(<\format Y_j,\format G_j> + \frac{\theta}{2}||\format G_j||^2_F\right),
\end{eqnarray}
where $<\cdot, \cdot>$ denotes the inner product, $\bf{Y}$ defines the Lagrangian multipliers (in matrix form), and $||\cdot||_F$ refers to the Frobenius norm that calculates the root of all the square entries' sum in a matrix. Here $\theta$ is a penalty parameter balancing the four equation constraints in Eq.~(\ref{eqs:standard model_2}), and follows the standard ALM updating rule as $\theta^{(k+1)}=\min\left(\rho \theta^{(k)},\theta_{max}\right)$, where $\rho$ and $\theta_{max}$ are both user-defined parameters, and $k$ indexes the iteration. The updating rules of the other variables including $\format S, \format L, \format N, \varepsilon$ and $\format Y$ are derived as follows.

{\vspace{3mm} \noindent{\bf Optimize $\format S$.~~~~}} By removing all the items irrelevant to $\format S_{1}$ in $f$, we can get
\begin{eqnarray*}
f(\format S_{1}) 
& = & ||\format S_{1}||_* + \frac{\theta}{2}||\format S_{1} - (\format L^{(k)} - \theta^{-1}\format Y_{1}^{(k)})||_F^2.
\end{eqnarray*}

According to the ALM algorithm, we can get the updating rule of $\format S_{1}$ as
\begin{equation}
\format S_{1}^{(k+1)} = Us_{\theta^{-1}}(S_{temp})V^T,
\label{eqs:update_S1}
\end{equation}
where $US_{temp}V^T$ is the SVD of $\format L^{(k)} - \theta^{-1}\format Y_{1}^{(k)}$, and
\begin{eqnarray*}
s_{\theta^{-1}}(x) = \begin{cases}
x - \theta^{-1}, & x > \theta^{-1}\\
x + \theta^{-1}, & x < -\theta^{-1}\\
0,& others.
\end{cases}
\end{eqnarray*}

Similarly, keeping only the items related to $\format S_{2}$ in $f$ yields
\begin{align*}
f(\format S_{2})
&= \lambda\left(||\format S_{2}||_1  + \frac{\theta}{2\lambda}||\format S_{2} - (\format P\format L^{(k)} - \theta^{-1} \format Y_{2}^{(k)})||_F^2\right),
\end{align*}
and we can get the updating rule of $\format S_2$ as
\begin{equation}
\format S_{2}^{(k+1)} = s_{\frac{\lambda}{\theta}}(\format P\format L^{(k)} - \theta^{-1}\format Y_{2}^{(k)})
\label{eqs:update_S2}.
\end{equation}

{\vspace{3mm} \noindent{\bf Optimize $\format L$ and $\format N$.~~~~}}By keeping only the items related to $\format L$, $f$ is simplified as
\begin{eqnarray*}
f(\format L)
=\frac{\theta}{2}||\format S_{1} - \format L + \theta^{-1}\FORMAT Y_{1}||_F^2  + \frac{\theta}{2}||\format S_{2} - \format P\format L + \theta^{-1}\FORMAT Y_{2}||_F^2 + \frac{\theta}{2}||\format M - \format L - \format N + \theta^{-1}\FORMAT Y_{3}||_F^2,
\end{eqnarray*}
and the partial derivative of $f(\format L)$ with respect to $\format L$ is
\begin{eqnarray*}
\frac{\partial f(\format L)}{\partial \format L} = \theta(\format L\!-\!\format S_{1}^{(k)}\!-\!\theta^{-1}\FORMAT Y_{1}^{(k)})\!+\!\theta[\format P^T\format P\format L\!-\!\format P^T(\format S_{2}^{(k)}\!+\!\theta^{-1}\FORMAT Y_{2}^{(k)})]
\!+\!\theta(\format L\!-\!\format M\!+\!\format N^{(k)}\!-\!\theta^{-1}\FORMAT Y_{3}^{(k)}).
\end{eqnarray*}

Similarly, the partial derivative of $f(\format N)$ with respect to $\format N$ is
\begin{eqnarray*}
\frac{\partial f(\format N)}{\partial \format N} = \theta(\format N - \format M + \format L^{(k)} - \theta^{-1}\FORMAT Y_{3}^{(k)}) + \theta[\format N\odot \format N - 9 \bm{\sigma}\odot \bm{\sigma} + \bm{\varepsilon}^{(k)} + \theta^{-1}\FORMAT Y_{4}^{(k)}]\odot 2\format N.
\end{eqnarray*}

Obviously it is hard to get the closed-form solution to either $\frac{\partial f(\format L)}{\partial \format L} = 0$ or $\frac{\partial f(\format N)}{\partial \format N} = 0$, so we resort to the gradient descent method to approximatively update these two variables as
\begin{eqnarray}
\format L^{(k+1)} = \format L^{(k)} - \Delta \times \frac{\partial f(\format L)}{\partial \format L}|_{\format L = \format L^{(k)}},
\label{eqs:update_L}\\
\format N^{(k+1)} = \format N^{(k)} - \Delta \times \frac{\partial f(\format N)}{\partial \format N}|_{\format N = \format N^{(k)}}.
\label{eqs:update_N}
\end{eqnarray}
Here $\Delta$ is the learning rate.

{\vspace{3mm} \noindent{\bf Optimize ${\bm{\varepsilon}}$.~~~~}}
The derivative of $f$ with respect to $\varepsilon$ is
\begin{eqnarray*}
\frac{\partial f(\bm{\varepsilon})}{\partial \bm{\varepsilon}} = \theta(\bm{\varepsilon} + \format N^{(k)}\odot \format N^{(k)} - 9 \bm{\sigma}\odot \bm{\sigma} + \theta^{-1}\FORMAT Y_{4}^{(k)}).
\end{eqnarray*}
Under the nonnegative assumption on ${\bm{\varepsilon}}$, we can get its updating rule as
\begin{eqnarray}
\bm{\varepsilon}^{(k+1)} = \max \left(9 \bm{\sigma}\odot \bm{\sigma} - \format N^{(k)}\odot \format N^{(k)} - \theta^{-1}\FORMAT Y_{4}^{(k)},~\format 0\right).
\label{eqs:update_epsilon}
\end{eqnarray}

All the algorithm parameters are set as follows: $\lambda$=0.2, $\theta^{(0)} = 1e-2$, $\rho = 1.6$, $\theta_{max} = 1e1$, $\Delta_{\format L} = 1e-2$ and $\Delta_{\format N} = 5e-2$. These constant parameters are set empirically by testing the algorithm on a series of OCT data to obtain the best denoising performance and least running time. Besides, all of the parameters are fixed across all the experiments in this paper. We take the averaged pre-registered frames as the initialization of the denoised frame.
For more clarity, the entire iterative algorithm based on the above derivations is summarized in Alg.~\ref{alg:standard}.
\begin{algorithm}[ht]
\SetKwInOut{Majorization}{Majorization}\SetKwInOut{Minimization}{Minimization}
\SetKwData{set}{set}
\SetKwInOut{Initialization}{Initialization}\SetKwInOut{Input}{Input}\SetKwInOut{Output}{Output}
\vspace{2mm}
\Input{Capturing data $\format M$ and estimated noise standard deviation $\bm{\sigma}$.}
\Output{Denoised frames $\format L$ and separated noise $\format N$.}
\vspace{2mm}
 $\format{L}^{(0)}=\overline{\format{M}}$, $\format{N}^{(0)}=\format{M}-\format{L}^{(0)}$,
 $\bm {\varepsilon}^{(0)}=\textbf{0}$; $\FORMAT Y_{1\cdots 4}^{(0)}=\textbf{0}$\;
      \While{not converged}{
          Update $\format S_{\{1,2\}}^{(k+1)} $ according to (\ref{eqs:update_S1}) and (\ref{eqs:update_S2})\;
          Update $\format{L}^{(k+1)}$ according to (\ref{eqs:update_L})\;
           Update $\format{N}^{(k+1)}$ according to (\ref{eqs:update_N})\;
          Update slack variables $ \bm{\varepsilon}^{(k+1)}$ according to (\ref{eqs:update_epsilon})\;
          $ \FORMAT Y_{\{1\dots4\}}^{(k+1)} = \FORMAT Y_{\{1\dots4\}}^{(k)} + \theta^{-1}\format G_{\{1\dots4\}}$\;
          $ \theta^{(k+1)}=\min\left(\rho \theta^{(k)},\theta_{max}\right)$\;
          $k:=k+1$.
          }
\caption{\small{The proposed multi-frame algorithm for OCT denoising}}
\label{alg:standard}
\end{algorithm}

\subsection{Evaluation criterion}
To quantitatively evaluate the denoising performance of different approaches, we utilize three widely used image quality criteria, including the peak signal-to-noise ratio (PSNR), the structure similarity (SSIM) \cite{SSIM} and the figure of merit (FOM) \cite{FOM_1, FOM_2} as evaluation metrics for the final denosing results.

\vspace{-2mm}
\paragraph{\bf PSNR.} In digital image recovery, PSNR has traditionally been widely used to assess the image quality of processed image $\format L_{m\times n}$, with respect to its ground truth $\format I_{m\times n}$. PSNR is calculated as
\begin{eqnarray}\label{eqs:PSNR}
PSNR = 10\times \log_{10}\left(\frac{MAX^2}{\frac{1}{mn}\sum_{i=1}^{m}\sum_{j=1}^{n}[\format L(i,j) - \format I(i,j)]^2}\right),
\end{eqnarray}
where $MAX = 2^b - 1$ is the maximum intensity of $b$ bit images. For example, for the widely used 8 bit images, $MAX = 255$. From the equation we can see that PSNR intuitively describes the intensity difference between two images, and would be smaller for low quality recovered images. Empirically, typical PSNR for visually promising images is roughly between 25 dB and 40 dB.

\vspace{-2mm}
\paragraph{\bf SSIM.} The structure similarity criterion is proposed in \cite{SSIM} to measure the structural similarities between two images. This criterion first selects two corresponding patch sets $\format p_{\format L}=\{p_{\format L}^k; k = 1\cdots K\}$ and $\format p_{\format I}=\{p_{\format I}^k; k = 1\cdots K\}$ from $\format L$ and $\format I$ respectively, with $K$ being the patch number, and then calculates the preliminary SSIM between each patch pair $p_{\format L}^k$ and $p_{\format I}^k$ as
\begin{eqnarray}\label{eqs:SSIM}
SSIM(p_{\format L}^k,p_{\format I}^k) = \frac{(2\mu_{\format L}^k\mu_{\format I}^k + c_1)(2\sigma_\format{L,I}^k + c_2)}{[(\mu_{\format L}^k)^2 + (\mu_{\format I}^k)^2 + c_1][(\sigma_{\format L}^k)^2 + (\sigma_{\format I}^k)^2 + c_2]},
\end{eqnarray}
where $\mu_{\format L}^k$ and $\mu_{\format I}^k$ are respectively the average pixel intensities of patch $p_{\format L}^k$ and $p_{\format I}^k$. $\sigma_{\format L}$ and $\sigma_{\format I}$ are the patchs' standard variances, while $\sigma_{\format{L,I}}$ is the covariance between $p_{\format L}^k$ and $p_{\format I}^k$. Besides, $c_1 = (k_1MAX)^2,c_2 = (k_2MAX)^2$ are two constants, with $k_1$ and $k_2$ being two use-defined parameters whose default values are respectively 0.01 and 0.03. The final SSIM score between two images is the average of all the patches' preliminary SSIM scores. The SSIM score ranges from 0 to 1, and is higher when two images own more similar structural information. Compared to traditional metrics such as PSNR, which only reveals the intensity differences between two images, SSIM reflects the similarity in structural information of an image pair, and thus is closer to human perception.

\vspace{-2mm}
\paragraph{\bf Edge preservation.} Further processing of denoised OCT images would likely involve the segmentation of layers or identification of a particular image feature. Thus the preservation of edges in denoised OCT images is very important. Here we also adopt the figure of merit (FOM) \cite{FOM_1, FOM_2} to evaluate the edge preservation abilities of various denoising methods. FOM is defined as
\begin{eqnarray}\label{eqs:FOM}
FOM = \frac{1}{max\left(n_{\format L},n_{\format I}\right)}\sum_{i=1}^{n_{\format L}}\frac{1}{1+\gamma d_i^2},
\end{eqnarray}
where $n_{\format L}$ and $n_{\format I}$ are respectively the numbers of detected edge pixels in the reconstructed image and the groundtruth image, $d_i$ is the Euclidean distance between the $i$th detected reconstructed edge pixel and its nearest groundtruth edge pixel, and $\gamma$ is a constant parameter typically set to be $\frac{1}{9}$. In this paper we use the Canny edge detector under default parameter settings in Matlab. FOM score ranges from 0 to 1, and is higher when the reconstructed image owns clearer edges and thus is more similar to the groundtruth image.

\section{Experimental results}\label{sec:Experimental Results}
In this section, we test our denoising approach on two public OCT image sets, and compare our denosing results with those of several previously reported popular methods. Besides the visual comparison, we also perform quantitative comparison in terms of PSNR, SSIM, FOM and running time.

\subsection{Results on the pigeye data and performance comparison}
In this experiment, we use the public pigeye OCT dataset in \cite{Data_Website}, which is also used as the source data in \cite{Multiframe_Wavelet}. The dataset is acquired using a Spectralis HRA $\&$ OCT (Heidelberg Engineering) to scan a pig eye in the high speed mode with 768 A-scans. There are totally 455 images (each frame contains 768$\times$496 pixels) in the dataset including 35 sets, and 13 frames are included in each set sharing the same imaging position. All the 35 positions correspond to a complete 0.384mm shift in the transversal direction. For each frame, the pixel spacing is 3.87$\mu$m in the axial direction, and 14$\mu$m in the transversal direction.
To assess the quality of the recovered images, we need a noise-free benchmark image for reference.
However, due to the high imaging speed, the captured images' SNR is very low. Therefore, we utilize the same technique in \cite{Multiframe_Wavelet}, in which the averaged image of all the 455 pre-registered frames is used as the latent noise free image.

\begin{figure*}[t]
\centering
\includegraphics[width=0.6\textwidth]{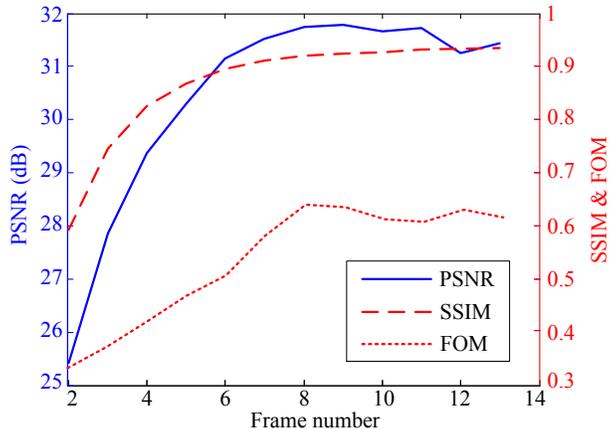}
  \caption{{\bf Reconstruction quality vs. input frame number.} Input: 2-14 frames of the pigeye data. The solid blue line corresponds to the axis on the left ranging from 25 to 32, while the two dashed red lines correspond to the axis on the right ranging from 0.3 to 1.}
  \label{fig:Frames} 
\end{figure*}

Since the proposed method is multi-frame based, we first investigate the effect of the most important parameter---the number of input frames---on our algorithm. Fixing all the other parameters, we run a Matlab implementation of our proposed method on an Intel E7500 2.93 GHz CPU computer with 4GB RAM and 64 bit Windows 7 system, and compare the performance of different numbers of input frames, ranging from 2 to 13. From the result in Fig. \ref{fig:Frames}, we can see that our algorithm still works using only 2 frames, and the reconstruction quality gradually improves using increasing number of frames from 2 to 8. This trend is much less obvious with more than 8 frames. Based on this observation, we use 8 input frames in the following experiments to compare our approach with other methods.

\begin{figure*}[!ht]
\centering
\includegraphics[width=1.0\textwidth]{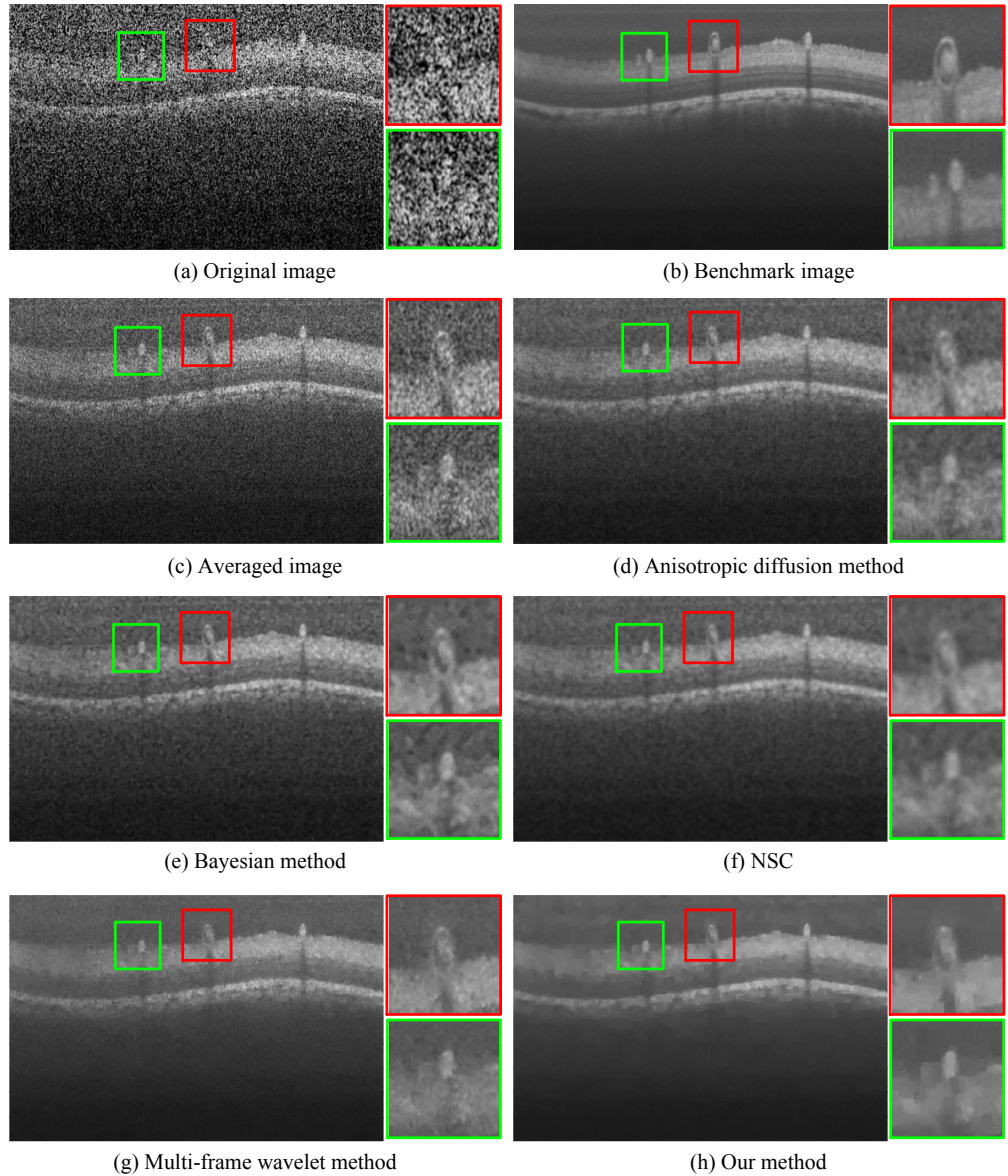}
  \caption{{\bf Comparison with the other four popular methods.} Input: 8 frames of the pigeye data. (a) is the original image in log transformed space, while (b) is the averaged image of 455 registered frames. (c) is the averaged image of the input 8 frames, and (d)-(g) are the recovered results of four popular methods. The result of our method is shown in (h). The two clipped patches on the right of each subfigure are closeups of the regions of interest.}
  \label{fig:STAR} 
\end{figure*}

Next, we run our algorithm and the other four popular denoising methods on the OCT dataset for comparison. The four methods include the complex diffusion method \cite{Diffusion}, the Bayesian method \cite{Bayesian}, NSC \cite{Noise_Estimation} and the multi-frame wavelet OCT denoising method \cite{Multiframe_Wavelet}. To validate the superior effectiveness of our approach, we compare all the algorithms' performance visually and quantitatively. What should be noticed is that the complex diffusion method, the Bayesian method and NSC are all single-frame methods, so we take the average image of the registered input frames as their single-frame input. Besides, the first two methods assume spatially invariant noise parameter (standard deviation). Correspondingly, we use the maximum in the estimated deviation matrix as the input standard deviation of noise. By random selection, the serial numbers of the input 8 sequential frames are from '$35\_6$' to '$35\_13$'.

The recovered images are shown in Fig.~\ref{fig:STAR}, where only the first frame is presented for each method. 
We can see that the recovered images of the anisotropic diffusion method, the Bayesian method and NSC still contain undesired noise, which largely degenerates the image quality and makes these three methods less competitive to the other two techniques.
On the whole, the multi-frame wavelet method and our method are both superior to the three single-frame approaches. Comparing the results of these two multi-frame methods, we can see that in the smooth regions, the wavelet method leaves out more noise than our method. In the textured regions, the result of our method maintains higher color contrast which would improve the visual effects. Stronger comparison is presented in the closeups. For example, in the green-rectangle-highlighted region, the wavelet method nearly blurs out the details of the white spot on the left side, while our method still contains grey value changes which would provide important information for diagnosis.

\begin{table*}[t]
\centering
\caption{Quantitative comparisons among different denoising methods}\label{tab:PSNR_SSIM}
{\small
\begin{tabular}{p{0.08\linewidth}p{0.16\linewidth}p{0.07\linewidth}p{0.07\linewidth}p{0.08\linewidth}p{0.08\linewidth}p{0.06\linewidth}p{0.07\linewidth}p{0.06\linewidth}}
\cline{1-9}
&Metric& Input& Average& Diffusion & Bayesian & NSC* & Wavelet & Ours\\
\cline{1-9}
Entire&PSNR(dB) & 17.19    & 24.56     &  29.14    & 28.38 & 29.82& 30.75 & {\bf 31.74}\\
\cline{2-9}
Image&SSIM     & 0.12   & 0.45    & 0.73  & 0.70  & 0.81 & 0.86 & {\bf 0.91}\\
\cline{2-9}
&Running time(s)&---& --- & 79 & 33 & {\bf 2} & 60 & 36\\
\cline{1-9}
Red&PSNR(dB) &15.03 & 22.02  & 26.60  & 26.07 & 27.47 & 27.85 & {\bf 28.92}\\
\cline{2-9}
Clip&SSIM     & 0.06 & 0.29 & 0.65 & 0.63 & 0.71 & 0.73 & {\bf 0.81}\\
\cline{2-9}
&FOM          & 0.43 & 0.46 & 0.51 & 0.57 & 0.60 & 0.61 & {\bf 0.63}\\
\cline{1-9}
Green&PSNR(dB) &15.13 & 21.91  & 26.14  & 25.35 &  26.83 & 27.83 & {\bf 28.75}\\
\cline{2-9}
Clip&SSIM     & 0.06 & 0.25 & 0.60 & 0.57 & 0.66 & 0.72 & {\bf 0.80}\\
\cline{2-9}
&FOM          & 0.48 & 0.49 & 0.51 & 0.53 & 0.58 & 0.58 & {\bf 0.58}\\
\cline{1-9}
\end{tabular}
}
\begin{tablenotes}
\footnotesize
\item[*] $^*$ The performance of NSC is tested by its proposers on the AMD Athlon X3 II CPU with 8GB of RAM and Windows 7 64-bit system, using Matlab and C++ programming for high computation efficiency.
\end{tablenotes}
\end{table*}

Numerical assessments are shown in Tab.~\ref{tab:PSNR_SSIM}. For the entire image, we can see that our method could raise the noisy images's PSNR from 17.19dB to 31.74dB, and SSIM from 0.13 to 0.91. In terms of all the three evaluation criteria including PSNR, SSIM and FOM, our method consistently owns advantages over the other methods.
Comparing the two multi-frame methods, namely the multi-frame wavelet method and our method, we can see that our approach is superior in PSNR and SSIM by around 1dB and 0.1, respectively. Our superior performance is mainly attributed to two factors: the combinational constraints from both the inter-frame and intra-frame priors, and the good convergence of the derived algorithm. Comparing the numerical evaluation results of the two selected regions of interests, we can also see clearer advantages of our method over the other methods.
What's more, the running time comparison is also provided in Tab.~\ref{tab:PSNR_SSIM} (the noise estimation time is also included for all the four algorithms). We run the Matlab codes of our algorithm and the other three methods except for NSC on our computer, while the running time of NSC is provided by its proposers who run their Matlab $\&$ C++ implementation on a different platform. We can see that our approach needs around 36s to process one frame, and is of similar efficiency to the Bayesian method which is the fastest algorithm among the four popular methods except for NSC.

\subsection{Analyzing OCT images of human subjects}

To further test the practical denoising effectiveness of our method, we conduct a denoising experiment on human retinal OCT images. We use the same public dataset as that used in \cite{Data_Human}, which is acquired by a SDOCT imaging system from Bioptigen Inc. with $\sim$4.5$\mu$m axial resolution, 500 A-scans per B-scan and 5 azimuthally repeated B-scans in each volume. Similar to the processing progress described in Sec. \ref{sec:NoiseEstimation}, we firstly register the OCT frames and then use different methods to denoise these frames. Considering that the anisotropic diffusion method, the Bayesian method and NSC leave too much noise on the recovered images, and thus own little competitiveness compared to the other two multi-frame methods, here we only present the denoising results of the multi-frame wavelet method and our method for clearer comparison.

\begin{figure*}[t]
\centering
\includegraphics[width=0.9\textwidth]{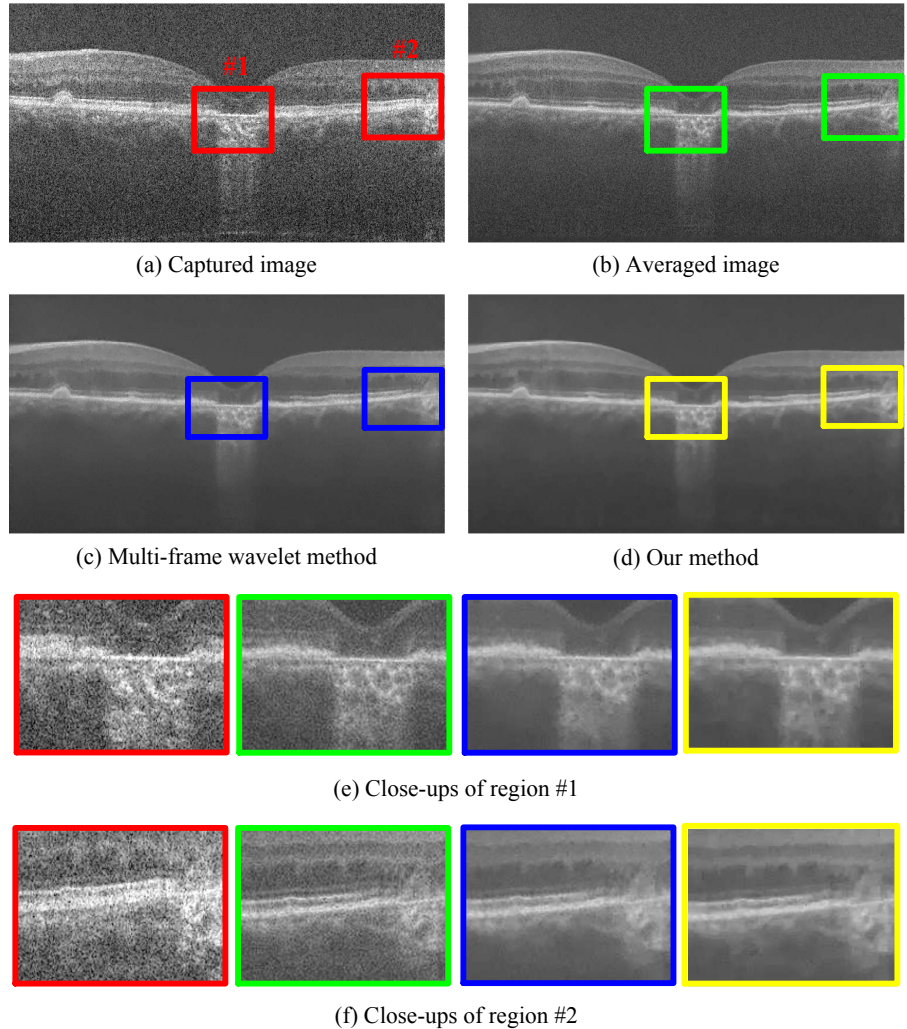}
  \caption{{\bf Denoising results of the human retinal OCT images.} Input: 5 frames of the human retina data. (a) shows one of the 5 captured frames, and (b) is the average of the 5 frames. (c) and (d) are respectively the results of the multi-frame wavelet method and our method. Close-ups of two selected regions of interest are shown in (e) and (f), which offer a clearer comparison.}
  \label{fig:Human} 
\end{figure*}

The results are shown in Fig. \ref{fig:Human}, which exhibits similar performance ranking to that of the pigeye data.
Comparing the denoising results produced by the multi-frame wavelet method and our proposed method carefully, we can see that the result of the wavelet method contains undesired edge burrs, while our result presents clearer layer boundaries (such as the horizontal layer edges in the two selected regions), which would help a lot in follow-up analysis of the denoised images, such as OCT layer segmentation and diagnosis.



\section{Conclusions and discussions}\label{sec:Conclusions and discussions}

\subsection{Summary and conclusions}

In this study, we propose a multi-frame OCT denoising method utilizing constraints from both inter-frame and intra-frame priors. Specifically, the inter-frame prior refers to the low rank of registered OCT frames, and the intra-frame prior is the sparsity of image gradient. Benefited from the proper convexification transformations and usage of ALM, the derived algorithm converges well on different data.
Besides, by incorporating a non-parametric and non-uniform noise description, our approach is applicable for different noise models.

On the adopted benchmark data, our approach could improve OCT image's quality by raising the image's PSNR from 17.19dB to 31.74dB and SSIM from 0.12 to 0.91, in around 36s for each frame. The comparisons with the other four popular methods on both pigeye data and human retinal data reveal that our method owns advantages mainly in two aspects:
(1) being able to attenuate speckle noise effectively and preserve crucial image details; (2) with efficiency comparable to the reported fastest approaches.
Such high performance of the proposed method mainly benefits from the combinational prior modeling and effective optimization algorithm.

\subsection{Limitations and future extensions}

The performance of our method depends on the registration accuracy, due to that the low rank prior in the objective function does not hold for an unaligned frame stack. This is also a challenge for other multi-frame denoising methods, and need to be addressed by the progress of noise robust matching techniques. Besides, a larger frame number is favourable to take advantage of the low rank prior. Therefore, one need to set the system's frame rate to balance the noise level and the number of available frames in practical applications.

In addition, the widely used anisotropic total variation is utilized as the intra-frame prior, which penalizes the diagonal gradients more significantly than the horizontal and vertical ones. This means that the utilized non-uniform constraint on image intensity changes along different directions, which may introduce undesired artifacts in the denoised images. 
Thus exploring an isotropic intra-frame prior would be one of our future extensions.


\end{document}